# Research on an intelligent fault diagnosis method for nuclear power plants based on ETCN-SSA combined algorithm

Jiayan Fang [a], Siwei Li [a], Yichun Wu [a, *], Ming He [b], Fengtao Xu [b]

[a] *College of Energy, Xiamen University, Xiamen, Fujian 361102, China*

[b] *Fujian Ningde Nuclear Power Co., Ltd., Ningde Fujian 355200, China*

*\* Corresponding Author:*

*Yichun Wu; College of Energy, Xiamen University; Xiang'an Nan Road 4221, Xiang'an District, Xiamen, Fujian 361102, China; ycwu@xmu.edu.cn; 086-152-8061-3500.*

*E-mail: ycwu@xmu.edu.cn*

**Abstract:** Utilizing fault diagnosis methods is crucial for nuclear power professionals to achieve efficient and accurate fault diagnosis for nuclear power plants (NPPs). The performance of traditional methods is limited by their dependence on complex feature extraction and skilled expert knowledge, which can be time-consuming and subjective. This paper proposes a novel intelligent fault diagnosis method for NPPs that combines enhanced temporal convolutional network (ETCN) with sparrow search algorithm (SSA). ETCN utilizes temporal convolutional network (TCN), self-attention (SA) mechanism and residual block for enhancing performance. ETCN excels at extracting local features and capturing time series information, while SSA adaptively optimizes its hyperparameters for superior performance. The proposed method's performance is experimentally verified on a CPR1000 simulation dataset. Compared to other advanced intelligent fault diagnosis methods, the proposed one demonstrates superior performance across all evaluation metrics. This makes it a promising tool for NPP intelligent fault diagnosis, ultimately enhancing operational reliability.

**Nomenclature**

AI: Artificial Intelligence

BN: Batch Normalization

CNN: Convolutional Neural Network

Conv: Convolutional

ETCN: Enhanced Temporal Convolutional Network

FDD: Fault Detection and Diagnosis

GRU: Gated Recurrent Unit

I&C: Instrumentation and Control

LOCA: Loss of Coolant Accident

LSTM: Long Short-Term Memory

MSLB: Main Steam Line Break

NPPs: Nuclear Power Plants

RELU: Linear Rectification Function

ResNet: Residual Network

RNN: Recurrent Neural Network

SA: Self-attention

SG: Steam Generator

SGTR: Steam Generator Tube Rupture

SSA: Sparrow Search Algorithm

SVM: Support Vector Machine

TCN: Temporal Convolutional Network

t-SNE: T-Distributed Stochastic Neighbor Embedding

# 1 Introduction

Nuclear energy is emerging as a critical driver for decarbonization and fostering a greener global economy (Jamil et al., 2016). Given the large size, variable operating conditions, intricate coupling within NPPs, and the risk of radioactive hazards, the implementation of stringent safety methods is paramount to guarantee operational safety and reliability (Tolo et al., 2019). Besides, the potential for unforeseen failures in critical systems that cannot be fully anticipated or prevented can trigger abnormal events in NPPs, which not only disrupt stable operation but also incur unnecessary maintenance costs (Moshkbar-Bakhshayesh and Ghofrani, 2013). Therefore, prompt and accurate fault diagnosis is crucial for maintaining the safe and reliable operation of NPPs. To safeguard the safety and economic operation of NPPs, digital instrumentation and control (I&C) system plays a critical role in continuously monitoring vital parameters like temperature, pressure, and coolant flow rate (Lin et al., 2022). Current methods for diagnosing NPP faults primarily rely on expert knowledge and operators' experience, which limits independent diagnosis. Although the existing fault detection and diagnosis (FDD) system in NPPs such as the real-time information monitoring system developed by China General Nuclear Power Engineering Co., Ltd. can provide some assistance, such system is still heavily dependent on expert knowledge. Moreover, there are a large number of alarm signals, which may overwhelm operators' ability to process it rapidly and accurately when a fault occurs (Choi et al., 1995). In recent years, the continuous advancements in I&C system and artificial intelligence (AI) algorithms have been enabled to improve the results of fault diagnosis in NPPs, which can better utilize the system's operating data and assist operators in making more accurate decisions. Furthermore, this method can also promote the intelligent operation and maintenance of NPPs, leading to significant cost reductions for remote areas such as offshore platforms and polar regions (Gomez-Fernandez et al., 2020).

Over the years, three primary fault diagnosis methods for complex structures have been developed: model-based methods, knowledge-based methods, and data-driven methods. Model-based methods need mathematical models equipped with strong assumptions and physical knowledge, which are usually unrealistic for intricate non-linear dynamic systems such as NPPs. Knowledge-based methods require the long-term accumulation of expert knowledge, which can be time-consuming and expensive. Furthermore, the sheer number and complexity of potential fault scenarios can make it challenging to create an exhaustive set of rules. Compared with methods above, data-driven methods can skillfully utilize advanced algorithms to achieve comprehensive analysis of non-linear relationships within a large amount of data and flexibly judge the system status without too much professional knowledge (Yin et al., 2023b). Therefore, since data-driven methods entered the field of FDD, it has become a hot topic. Data-driven methods can be broadly divided into two categories: conventional machine learning and neural networks. Conventional machine learning methods, including logistic regression (Wang et al., 2023), decision trees (Subramaniyan, 2024; Zio et al., 2009), support vector machines (Cao et al., 2022; Wang and Sun, 2023), random forests (Prasojo et al., 2023; Yu et al., 2021) and K-nearest neighbors (Li et al., 2022), have been widely used. However, they still require some degree of expert knowledge, and their fault diagnosis ability are restricted to specific application scenarios. Most importantly, they struggle to establish complex, non-linear relationships between faults and sensor measurements, making it difficult to meet the requirements of NPP fault diagnosis. As an efficient network structure, neural network methods possess remarkable automatic learning and feature extraction capabilities, effectively overcoming the shortcomings of traditional machine learning models. Besides, they can dig into complex data to extract the non-linear mapping relationships between multidimensional parameters and faults (C. Wang et al., 2022).

Wang (Wang et al., 2021) introduced a NPP fault diagnosis framework based on a convolutional gated recurrent unit network to capture the characteristics of time series and enhanced particle swarm optimization to optimize the hyperparameters of the network. P. Wang (P. Wang et al., 2022) specifically focused on long short-term memory (LSTM) network, demonstrating its effectiveness in developing a fault diagnosis method for small pressurized water reactors with high accuracy and robustness. Convolutional neural network (CNN) can automatically extract features from high-dimensional data, which has been used in fault diagnosis for NPP safety management (He et al., 2021). By evaluating the system of general convolutional and recurrent structures, Bai (Bai et al., 2018) introduced a novel architecture called TCN that incorporates causal convolution, dilated convolution, and residual structure. In order to enhance adaptive diagnosis of short-circuit faults in electric vehicles, Zhang (Zhang et al., 2024) proposed a method combining adaptive threshold algorithm and TCN. Li (Li et al., 2021) proposed a new chiller fault diagnosis method that integrates feature enhancement and TCN, resulting in enhanced accuracy and reduced model complexity by effectively capturing the intricate dynamic coupling characteristics among variables. To overcome the challenges of real-time diagnosis and algorithm autonomous optimization, Ai (Ai et al., 2021) proposed a novel fault diagnosis method for hypersonic air vehicle sensors based on TCN and strengthen elite genetic algorithm. However, neural network methods may fail to capture crucial features from complex data, which hinder their ability to accurately diagnose faults (Li et al., 2024). SA mechanism improves the learning ability by selectively adjusting the feature processing of networks and reducing the sample data to be processed (Vaswani et al., 2017). Fahim (Fahim et al., 2020) proposed SA-CNN for transmission line fault detection, which can extract features from time series images. Shen (Shen et al., 2023) developed an integrated method combining SA mechanism and TCN to address the challenges of limited fault data and data imbalance, thereby enhancing chiller fault diagnosis accuracy. To enable effective fault diagnosis under limited labeled data and variable operating conditions, Zhu (Zhu et al., 2024) proposed a self-supervised fault diagnosis method with temporal predictive and similarity contrast learning embedded with SA mechanism. Given the challenge of exploding/vanishing gradient problem during training deep neural networks, residual network (ResNet) incorporating residual blocks came into being (He et al., 2015). Yin (Yin et al., 2023a) proposed a ResNet-SVMs model for adaptive deep feature extraction and robust fault diagnosis on NPP rotating machinery. Focusing on enhancing the accuracy and robustness of fault diagnosis in NPP rotating machinery, Yin (Yin et al., 2023b) proposed an intelligent method integrating multi-sensor data with deep residual neural network. Huang (Huang et al., 2024) proposed a fault diagnosis framework that employs ResNet architecture for fault type classification and integrates autoencoder and LSTM for fault localization.

Neural network fault diagnosis methods involve numerous hyperparameters, and the final diagnosis results depend largely on the settings of these hyperparameters. Many studies still rely on manual debugging or transfer mature models from other fields, which are difficult to ensure that the hyperparameters generated by the two solutions are optimal. At present, the commonly used hyperparameter optimization method is random search. It boasts higher search efficiency by randomly sampling within a given range and allow the search to move in the direction of local or global optima. Mainstream random search includes genetic algorithm (Zou et al., 2023), ant colony optimization (Zhao et al., 2020), artificial fish swarm algorithm (Shen et al., 2011), simulated annealing (Dhagat and Jujjavarapu, 2021), particle swarm optimization (Khuwaileh et al., 2021) and SSA (T. Wang et al., 2022).

To improve the FDD system, this paper proposes a novel intelligent fault diagnosis method based on SSA and ETCN that integrates TCN, SA mechanism and residual block to comprehensively analyze and differentiate various accident scenarios making full use of both normal and abnormal operational data. The main contributions of this paper can be summarized as follows:

1. An in-depth qualitative and quantitative analysis of NPP intelligent fault diagnosis method are performed. A comparative evaluation with other methods is conducted to assess the feasibility and superiority of the proposed NPP intelligent fault diagnosis method.

2. ETCN is aimed to provide sufficient feature extraction function for fault diagnosis in NPP. TCN effectively captures temporal relationships within NPP data, while SA mechanism allows for spontaneous extraction of internal correlations within the same data. Additionally, residual block contributes its strength in adaptive feature extraction. To achieve optimal performance, SSA serves as an intelligent search tool for ETCN hyperparameter optimization, automatically exploring the hyperparameter space to identify the ideal configuration.

3. The step size and the width of sliding window are determined based on the physical characteristics of NPPs, followed by verification of the method's ability to diagnose faults under this sliding window.

The rest of the paper is structured as follows. Section 2 details the neural network method employed and introduces the proposed intelligent fault diagnosis framework. Section 3 describes the experimental dataset's source, the choice of sliding window and the selection process for hyperparameter optimization. Section 4 validates the effectiveness of the proposed method using dataset simulated by CPR1000 simulator. The conclusion and future work directions are given in the last section.

## 2 Methodology

Intelligent fault diagnosis methods in NPPs, while providing valuable operating status information and assisting operators in decision-making, face significant implementation hurdles in the transition from proposed technology to engineering applications. To bridge this gap, this paper proposes a method that integrates TCN, SA mechanism and residual block, along with SSA for adaptive hyperparameter optimization. As illustrated in Fig. 1, the proposed method comprises five main parts:

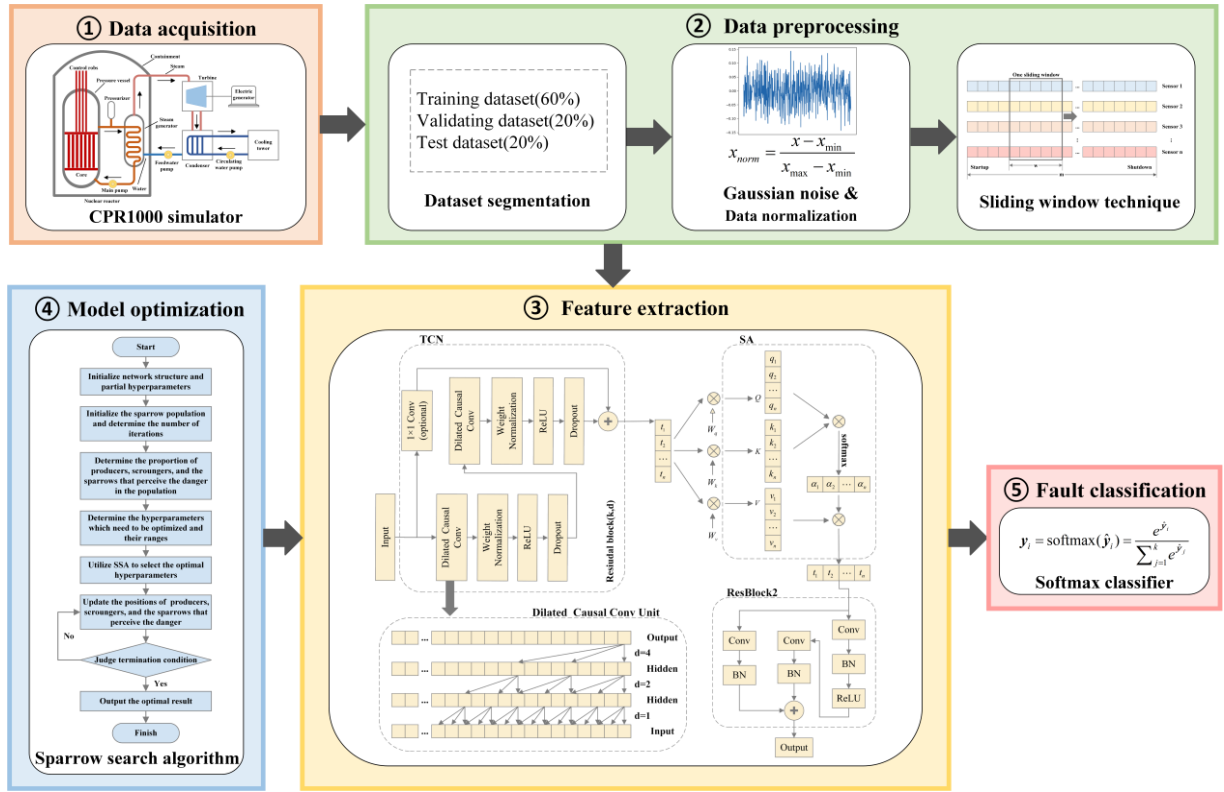

Fig. 1. General framework of ETCN-SAA intelligent fault diagnosis method.

(1) Data acquisition: The CPR1000 simulator based on the 3KEYMASTER simulation platform is used to simulate the corresponding operating conditions. By comparison and verification with the design manual, the data that matches the conditions is then obtained.

(2) Data preprocessing: NPP data is first split into training, verification and test dataset. Next, Gaussian noise is added to the data to improve model robustness, and then the data is normalized. Finally, the data is segmented into subsequences of appropriate length along the time axis through sliding window technique.

(3) Feature extraction: First, the dynamic time features in the multivariate time series data are learned through TCN. Then, by assigning weights to different parts of the data, SA mechanism can help the model focus on the most crucial features, which improves the model's understanding and utilization efficiency. Finally, residual block further extracts fault features from the processed data by adaptively capturing the relationships between features, enhancing its perception of different features.

(4) Fault classification: The softmax classifier is employed at the output layer to classify faults by assigning probabilities to each potential fault category.

(5) Model optimization: SSA is used to optimize ETCN's hyperparameters.

## 2.1 Data preprocessing

### 2.1.1 Gaussian noise

The quality of data significantly impacts the accuracy of the method. In practical applications, multivariate time series data is inherently noisy due to measurement instrument fluctuations. Therefore, to enhance the robustness of the proposed method and mimic the presence of real-world measurement noise, Gaussian noise is introduced to the original data. Each variable of multivariate time series data is overlapped with Gaussian noise with a normal distribution and 5% standard deviation.

### 2.1.2 Data normalization

In NPPs, sensor data often comes in different units and ranges. Generally, features with higher values tend to have a stronger influence on the learning process in neural networks. To address this, data normalization is necessary. Data normalization makes all features have a similar magnitude, thereby accelerating both the calculation speed and convergence speed. In the data preprocessing stage, the maximum-minimum normalization is used to linearly scale all variables to a common range, typically between 0 and 1. As shown in the Formula 1:

$$x_{norm} = \frac{x - x_{min}}{x_{max} - x_{min}} \tag{1}$$

where $x$ is the original value of the data, and $x_{norm}$ is the normalized value of the corresponding data. $x_{max}$ and $x_{min}$ are respectively the maximum and minimum values of the data.

### 2.1.3 Sliding window technique

NPP data are time series data, characterized by numerous operating parameters, rapid changes, and real-time demands. The analysis of single-point data cannot fully capture its complexity. Therefore, to fully grasp NPP behaviors, an intuitive approach is to use a two-dimensional sample matrix (feature variable dimension and time series dimension) for analyzing NPP operating status. This can be achieved through sliding window technique applied to the entire multi-sensor signal data. Sliding window technique, a common method for time series analysis, divides the normalized long time series into a set of subsequences along the time axis using a fixed time step to obtain the observation sequence that constitutes the model input. The window size is the same as the size of these subsequences, and its endpoint always represents the current moment. This ensures the capture of the latest data in the data stream, enabling the diagnostic model to identify changes in operating conditions and perform fault diagnosis more promptly.

Fig. 2 illustrates the generation of a two-dimensional matrix from a time series of size [n, m] using a sliding window with width w, assuming the system has n feature variables. The vertical axis represents the sensor variable dimension, while the horizontal axis represents the time series dimension.

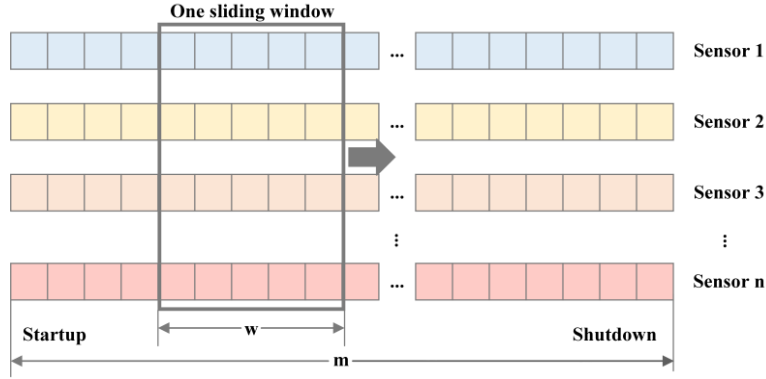

Fig. 2. Sliding window technique.

## 2.2 Feature extraction

### 2.2.1 Residual block

As the depth of CNN increases, its ability to extract richer feature information with enhanced non-linear expression capabilities improves. However, deeper networks can also experience performance degradation during training. ResNet is proposed to address this challenge, utilizing shortcut connections which help jump over layers in the network (He et al., 2015). These connections directly add the output of a previous layer to the input of a subsequent layer, effectively propagating the error from the low layer to the high layer and thereby alleviating the vanishing/exploding gradient problem (Gao et al., 2020). ResNet is usually composed of multiple residual blocks, each of which includes a convolutional (Conv) layer, a batch normalization (BN) layer, a non-linear activation function layer, and a shortcut connection.

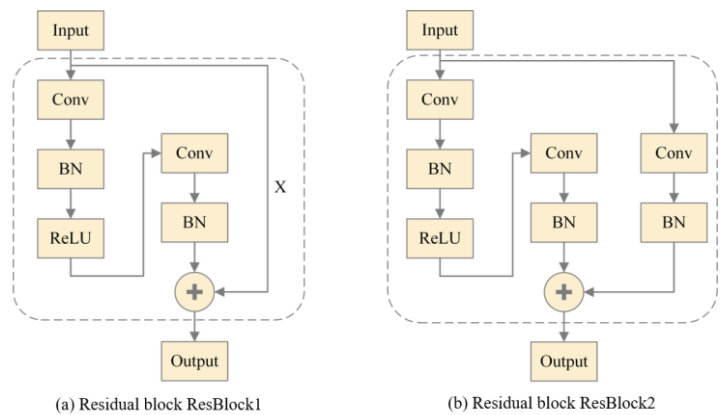

Fig. 3. Residual block structure.

Residual blocks come in two main types, as illustrated in Fig. 3. Both ResBlock1 and ResBlock2 include two Conv layers, two BN layers and a ReLU activation function. The key distinction lies in how they handle the shortcut connection. For input x, the output of the residual block can be expressed as:

$$y = f(F(x) + h(x)) \tag{2}$$

where $h(x)$ represents the mapping of shortcut connection. For ResBlock1, its shortcut connection is identity shortcut denoted by $h(x) = x$. For ResBlock2, its shortcut connection is convolution shortcut denoted by $h(x) = H(x)$, which includes a Conv layer and a BN layer. $F(x)$ represents the learned residual function.

This study will compare the effects of using ResBlock1 and ResBlock2 on improving model performance. So, we will construct two separate models, each containing only one residual block. One model will use ResBlock1 and the other will use ResBlock2. By comparing the performance of the two models, we can determine which type of residual block leads to better results in this specific case.

**2.2.2 Temporal convolutional network**

TCN is a sequence modeling method that uses time series as input, dilated causal convolutional layers as the basic architecture, and residual module as the connection. It aims to improve the ability of traditional CNN to extract the characteristics of time series (Bai et al., 2018). Compared with LSTM and other RNNs, TCN benefits from parallel processing capabilities, which allows for faster training and execution. The structure of TCN is illustrated in Fig. 4. It consists of a residual block which includes two weight normalization layers, two dilated causal convolution layers, two ReLU activation function layers, and two dropout layers. The one-dimensional full convolution layer ensures that the input and output sequences have the same length.

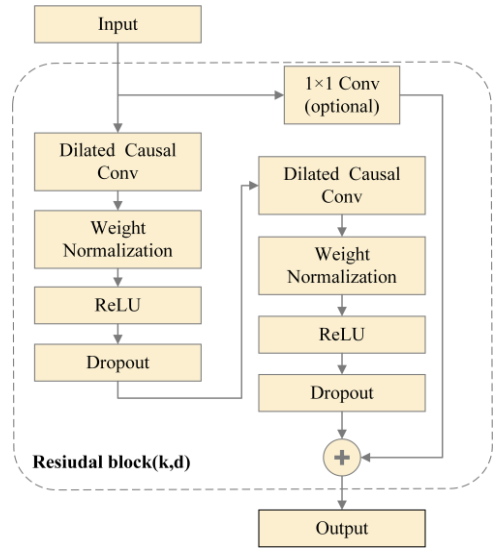

Fig. 4. TCN structure.

Causal convolution, a technique used in TCN, ensures the model only considers past information by applying left padding. This enforces time causality and avoids future data influence. However, the length of time series reviewed by causal convolution is limited by the size of convolution kernel and the depth of network structure.

Therefore, TCN uses dilated causal convolution. This technique changes the relationship between the receptive field range and network depth from linear to exponential, and thus significantly expands the model's ability to capture long-term dependencies while maintaining a shallow network structure. The difference between causal convolution and dilated causal convolution is shown in Fig. 5. The calculation formula for dilated causal convolution is shown in Formula 3:

$$F(s) = \sum_{i=0}^{k-1} f(i) \cdot x_{s-d \cdot i} \tag{3}$$

where $k$ represents the filter size, $d$ represents the dilation rate, $x$ represents the input sequence, $s$ represents the sequence element, $s - d \cdot i$ represents the past direction, and $f(i)$ represents a filter of size $i$ with $i \in [1, k-1]$. Dilation rate controls the stretching of the filter, allowing it to capture information from further past elements in the sequence. As the number of network layers increases, the effective receptive field grows exponentially due to dilation, which enables the network to learn long-term connections between data. When $d=1$, the dilated causal convolution becomes a causal convolution.

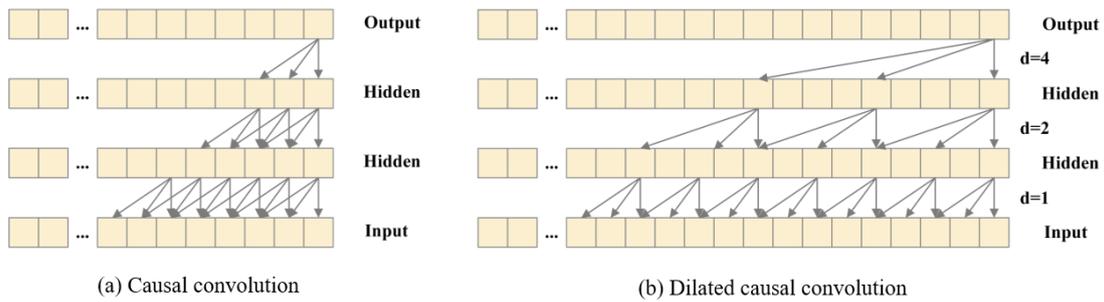

Fig. 5. Two convolution structures.

### 2.2.3 Self-attention mechanism

SA mechanism is widely used in natural language processing and computer vision. By mimicking human cognition in focusing on key features, SA mechanism assigns weights and distributes input to network nodes, effectively highlighting relevant information and suppressing irrelevant features. In the context of NPP intelligent fault diagnosis, this approach is particularly valuable. Sensor data from NPPs can be massive and high-dimensional, containing diverse physical quantities and characteristics. By applying SA mechanism, the model can exploit its parameter sharing capabilities to selectively focus on key information for analysis. This leads to several advantages: a significant reduction in the number of model parameters, a decrease in overall model complexity, and a lower risk of overfitting. Consequently, the model's generalization ability is enhanced, leading to more accurate fault diagnosis across various NPP operating conditions. SA mechanism is illustrated in Fig. 6.

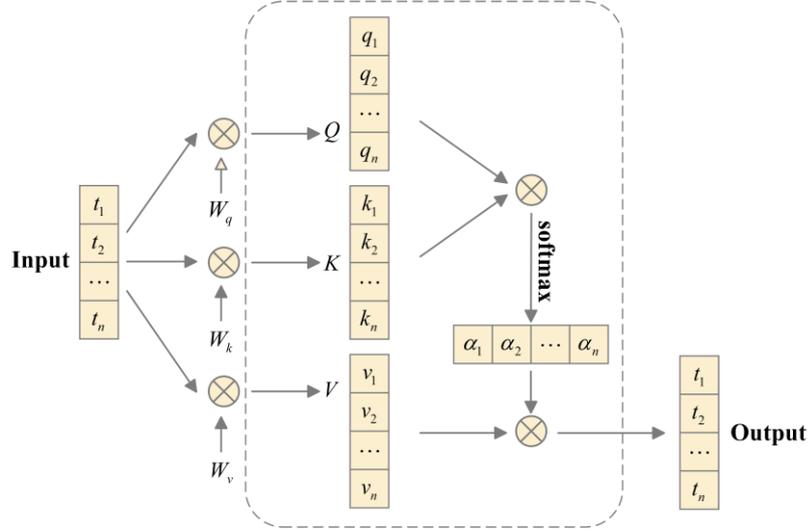

Fig. 6. SA mechanism structure.

SA mechanism operates in three key steps. First, the preprocessed data is transformed into three new vectors for each element: a query vector, a key vector, and a value vector. Then, the information between the query vector and the key vector is calculated and softmax is used to obtain the attention score $\alpha$. Finally, the contextual information is extracted based on the attention score. The steps involved are as follows:

$$Q = W_q \cdot I \tag{4}$$

$$K = W_k \cdot I \tag{5}$$

$$V = W_v \cdot I \tag{6}$$

$$S = \frac{Q \cdot K^T}{\sqrt{d_k}} \tag{7}$$

$$\alpha = \text{softmax}(S) \tag{8}$$

$$\text{output} = \alpha \cdot V \tag{9}$$

where $W_q$, $W_k$ and $W_v$ represent the weight matrices of $Q$, $K$ and $V$ respectively, $I$ represents the input matrix, $d_k$ represents the scaling rate, and softmax will be discussed in Section 2.3.1.

## 2.3 Classification layer and model optimization

### 2.3.1 Softmax

Softmax, shown in Formula 10, is used for post-processing after feature extraction and is valuable in various multi-class classification tasks.

$$y_i = \text{softmax}(\hat{y}_i) = \frac{e^{\hat{y}_i}}{\sum_{j=1}^{k} e^{\hat{y}_j}} \tag{10}$$

where $\hat{y}_i$ represents the vector fed to softmax layer, $k$ represents the number of output classes, and $y_i$ represents the probability that the event that is happening belongs to class $i$. Softmax ensures that the output value is between 0 and 1 and the sum is 1. Finally, the class with the highest probability is considered the most likely prediction.

**2.3.2 Sparrow search algorithm**

Efficiently identifying the optimal hyperparameters of a neural network can significantly improve the model building process. The optimization process involves searching for different configurations, each of which is evaluated using validation loss. Swarm intelligence optimization algorithm, a popular approach in computational intelligence, mimic the collective behavior of animal groups in nature. By simulating the interactions and cooperation within these groups, such as fish schools, bird flocks, and bee colonies, it can achieve optimal purpose through simple interactions between individuals.

SSA mainly simulates the predation and anti-predation behavior of sparrows. It represents potential hyperparameter combinations by sparrow positions with different fitness. This method exhibits good performance in search accuracy, convergence speed, and stability (Xue and Shen, 2020). By overcoming limitations of the search space and avoiding local optima, SSA also promotes further exploration of unknown territory. According to certain rules, sparrows are categorized into three groups: producers, scroungers, and the sparrows that perceive the danger. Producers guide the population by searching for food sources, while scroungers follow their lead. Sparrows can usually switch between these roles through individual competition, but also fight against other predators as a whole. These three groups will continuously update their positions and identities in each iteration. Given maximum iterations, the number of producers, the number of sparrows that perceive the danger, the total number of sparrows, and the safety threshold, the position with the lowest fitness represents the optimal hyperparameter combination after a finite number of iterations.

# 3 Case experiment

## 3.1 Dataset description

Due to the challenges of acquiring real fault data from NPPs, this study specifically employs a CPR1000 simulator, which is based on the 3KEYMASTER simulation platform. The simulation data includes normal and various fault scenarios. All of the simulation scenarios are consistently operated at 100% power level. Following 40 seconds of steady-state operation, four operating transients are inserted and the entire process lasts for 15 minutes. The four operating transients include one steady state and three design basis accidents whose approximate positions are shown in Fig. 7. The failure point in the simulation is designated as Loop1, exhibiting slight variations in its variables compared to those of Loop2 and Loop3. As shown in Table 1, each design basis accident includes 20 different rupture sizes and. Data sampling occurred every second, resulting in time series data for 26 monitoring variables, as shown in Table 2. To assess the effectiveness and robustness of the proposed intelligent fault diagnosis method for NPPs, the collected data is divided into training dataset, validation dataset, and test dataset with a 6:2:2 ratio.

All experiments are conducted on a server equipped with an Intel Xeon Platinum 8362 CPU (2.80GHz) and an NVIDIA RTX 3090 GPU. The proposed framework is based on Pytorch 2.0.0 in Python 3.8.

Table 1: Description of simulation operational scenarios.

| Operational Scenario | Abbreviation | Label | Severity | Count of Samples |
|---|---|---|---|---|
| Normal | NO | 0 | - | 3300 |
| Loss of coolant accident | LOCA | 1 | 2.5%-50% | 17200 |
| Main steam line break | MSLB | 2 | 2.5%-50% | 17200 |
| Steam generator tube rupture | SGTR | 3 | 2.5%-50% | 17200 |

Table 2: State variables

| ID | Description | Unit |
|---|---|---|
| 1 | 1#Hot-leg temperature | °C |
| 2 | 2#Hot-leg temperature | °C |
| 3 | 3#Hot-leg temperature | °C |
| 4 | 1#Cold-leg temperature | °C |
| 5 | 2#Cold-leg temperature | °C |
| 6 | 3#Cold-leg temperature | °C |
| 7 | 1#Coolant flow rate | t/h |
| 8 | 2#Coolant flow rate | t/h |
| 9 | 3#Coolant flow rate | t/h |
| 10 | 1#SG steam flow rate | t/h |
| 11 | 2#SG steam flow rate | t/h |
| 12 | 3#SG steam flow rate | t/h |
| 13 | 1#SG pressure | MPa |
| 14 | 2#SG pressure | MPa |
| 15 | 3#SG pressure | MPa |
| 16 | 1#SG level (wide range) | m |
| 17 | 2#SG level (wide range) | m |
| 18 | 3#SG level (wide range) | m |
| 19 | 1#SG level (narrow range) | m |
| 20 | 2#SG level (narrow range) | m |
| 21 | 3#SG level (narrow range) | m |
| 22 | Total power | MW |
| 23 | Reactor operating pressure | MPa |
| 24 | Reactor outlet temperature | °C |
| 25 | Pressurizer pressure | MPa |
| 26 | Pressurizer level | m |

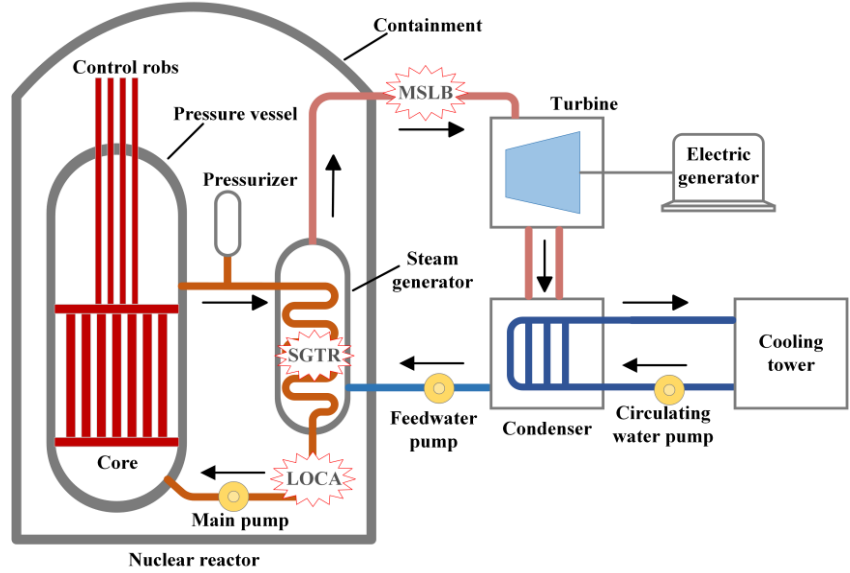

Fig. 7. Positions of three design basis accidents.

### 3.2 Model evaluation metrics

When evaluating classification models, various metrics are employed. These can be broadly categorized into two groups: overall classification performance like accuracy and specific recognition ability like precision, recall and F1-score.

Accuracy indicates the proportion of cases with correct diagnosis within the total samples. Precision, also known as positive predictive value, indicates the proportion of correctly identified healthy samples among all samples detected without faults. Recall indicates the proportion of actual fault-free samples that are correctly diagnosed. F1 score is a combined metric that considers both precision and recall. The closer the four values are to 1, the better the model fault diagnosis effect.

Their expressions are as follows:

$$Accuracy = \frac{TP+TN}{TP+FP+TN+FN} \tag{11}$$

$$Precision = \frac{TP}{TP+FP} \tag{12}$$

$$Recall = \frac{TP}{TP+FN} \tag{13}$$

$$F_1\_score = 2 \cdot \frac{Precision \cdot Recall}{Precision + Recall} \tag{14}$$

Where TP (True Positive) is a correctly positive result, TN (True Negative) is a correctly negative test result, FP (False Positive) is a false alarm, and FN (False Negative) is a missed case.

## 3.3 The choice of sliding window

The effect of sliding window to obtain data mainly depends on two key parameters: the step size used to move window across data and the width of sliding window. An appropriate step size determines the overlap between consecutive windows, while an appropriate window width can capture a sufficient amount of temporal data to effectively represent the evolving NPP operational state. A sliding step size that is too large can lead to rapid update of window data, potentially reducing sensitivity to subtle changes in the time series. Conversely, a very small step size can significantly increase the number of model iterations, leading to a slower update rate. Given the short duration of a typical nuclear reactor accident, a small step size is crucial for capturing rapid changes in NPP data. Therefore, we use the minimum step size of one second provided by the simulation model.

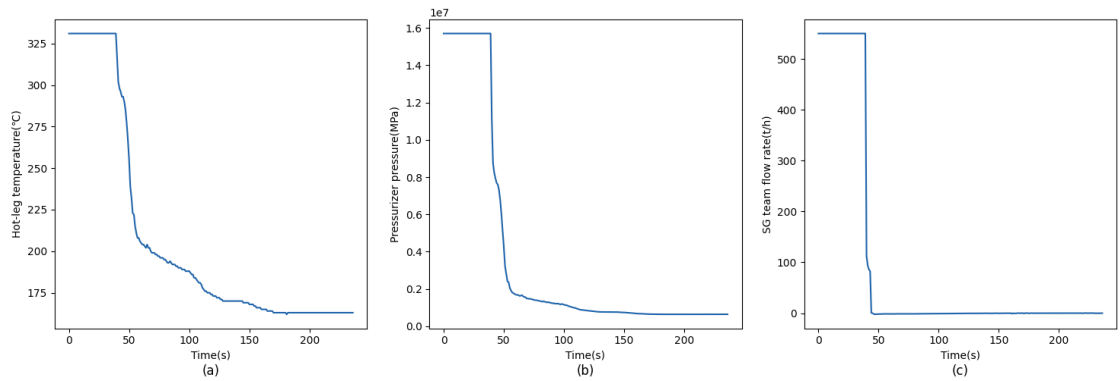

Fig. 8. Simulation results of (a) Hot-leg temperature, (b) Pressurizer pressure, (c) SG team flow rate.

The sliding window width can be determined according to the characteristics of system. While a larger sliding window width can capture more information which is conducive to extracting features hidden in the sequence, it leads to less data variation, hindering the diagnosis of accident conditions. However, an excessively small sliding window width may contain insufficient data, potentially hindering the model's ability to diagnose faults. Therefore, to determine the optimal sliding window width, the sliding window widths of 1, 2, 3, 4, and 5 minutes are selected. Taking LOCA accident as an example, Fig. 8 shows the simulation curves of hot-leg temperature, pressurizer pressure ang SG team flow rate. The corresponding calculation results for average, variance, and standard deviation with different sliding window width are presented in Fig. 9.

As illustrated in Fig. 9 (a), (d), and (g), the mean value of three variables is reflected. A large sliding window width results in a smoother trend, making it difficult to capture the rapid changes that might indicate an accident. Therefore, the sliding window width cannot be too large. As shown in the rest of Fig. 9, the variance and standard deviation of the data represent the overall fluctuations of the data. The small variance and standard deviation allow for a more accurate representation of the true changes in relevant variables. Since the time between the initial signs and the full-blown accident in NPPs is short, if the sliding window width is too large, the data with relatively stable changes before and after the accident may dilute the data with large fluctuations around the accident itself, which means that the window data will only reflect the long-term trend and miss the drastic parameter fluctuations To sum up, a large sliding window width tends to neglect the significant changes and fluctuations, reducing the diagnostic model's sensitivity to accident. When the sliding window width is too small, it may not contain enough data for fault diagnosis and can lead to misinterpretations due to similarities between normal and accident operating conditions at the initial phase. Therefore, a sliding window width of 2 minutes offers a balance between capturing the rapid changes during the symptom period and maintaining sufficient data redundancy for accurate fault detection.

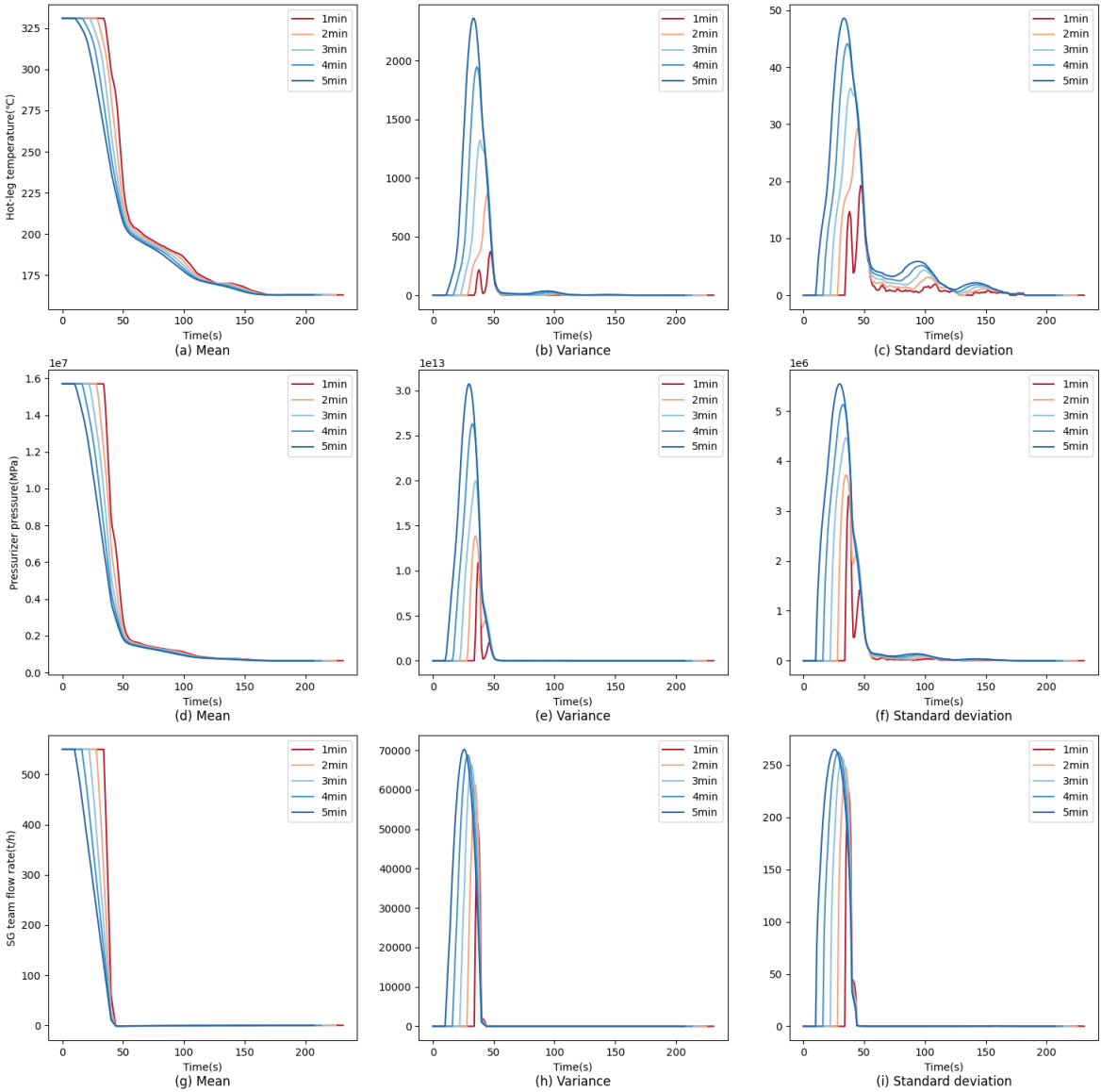

Fig. 9. Analysis results of sliding window width.

## 3.4 Model optimization

Neural networks have two types of parameters: weight parameters and hyperparameters. Weight parameters are learned during training and dynamically updated through back-propagation algorithm and optimizer. They determine the connection strength between neurons, thereby influencing the network's response and output. Adjusting these parameters can change the network's fitting ability, generalization ability and convergence speed. Hyperparameters are set before training and influence aspects like learning speed, convergence, and susceptibility to overfitting and underfitting. Manual parameter adjustment is not an ideal method because it is inefficient and prone to getting stuck in suboptimal configurations. Therefore, SSA is used to optimize these hyperparameters. The flowchart of SSA is shown in Fig. 10, and the specific steps are outlined as follows:

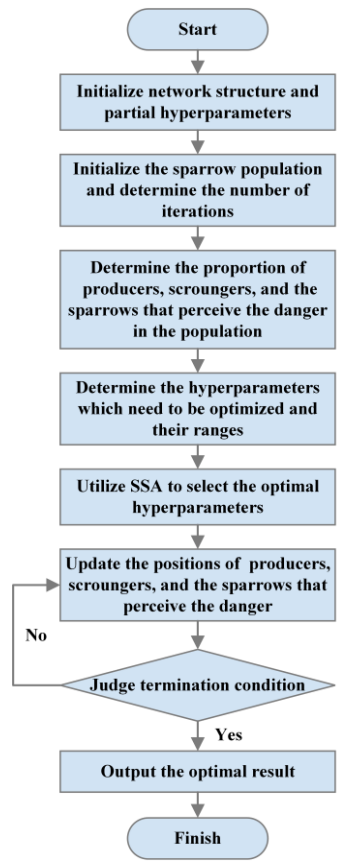

Fig. 10. Flowchart of SSA.

(1) Initialize the network structure and confirm some hyperparameters in advance.

(2) Initialize the parameters of SSA, such as population size, maximum number of iterations, safety threshold, warning threshold, and monitoring ratio.

(3) Determine the hyperparameters to be optimized and their search ranges.

(4) Execute SSA to calculate the fitness and rank them on their performance.

(5) Update the position of the sparrow population.

(6) Re-evaluate the fitness of the sparrows and update producers if a better solution is found.;

(7) Repeat step (5) and step (6) until a stopping criterion is met, and output the optimal result as the hyperparameters for the neural network;

(8) Train the neural network model using the obtained optimal hyperparameters and evaluate the model's performance.

The accuracy of fault diagnosis on the test data is used as the fitness value. The final values of specific hyperparameters are detailed in Table 3, while the remaining hyperparameters listed in Table 4 are determined in advance for reducing the number of variables and speeding up the optimization process. As presented in Table 3, the optimization results of SSA for ETCN hyperparameters are as follows: the learning rate is 0.000106, the kernel size is 5, the dropout rate is 0.289, the convolution kernel is 32, and the batch size is 128. The fitness value change process during the SSA optimization process is illustrated in Fig. 11.

Table 3: Selected hyperparameters needing optimization.

| Hyperparameter | Search range | Optimal value |
| --- | --- | --- |
| Learning rate | [0.001:0.000001:0.01] | 0.000106 |
| Kernel size | [3:1:12] | 5 |
| Dropout rate | [0.1:0.001:0.5] | 0.289 |
| Convolution kernel | [4,8,16,32,64] | 32 |
| Batch size | [32,64,128,256] | 128 |

[1]$[a:n:b]$ represents a discrete data set with the minimum value $a$, the maximum value $b$ and an interval $n$.

Table 4: Hyperparameters determined in advance.

| Hyperparameter | Value |
| --- | --- |
| Epochs | 100 |
| Optimizer | Adam |
| Activation | ReLU |
| Loss | Cross entropy loss |
| Kernel size of ResBlock1/ ResBlock1 | 3 |

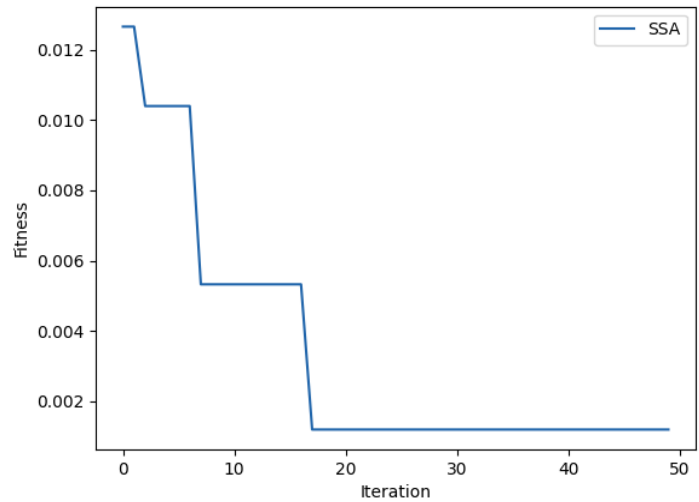

Fig. 11. Curve of SSA optimizing hyperparameters.

## 4 Model analysis

### 4.1 Model training and validation

There are four scenarios following the training process. The first is underfitting, which occurs when both training loss and validation loss are high. The second is overfitting, where training loss is low but validation loss is high. The third is good performance, where validation loss is low and close to training loss. The fourth possibility, where validation loss is low and training loss is high, is unlikely to occur because it suggests the model performs better on unseen data than the data trained, which goes against the principles of machine learning.

After obtaining the optimal hyperparameters in Section 3.4, ETCN is trained on training and validation dataset, verifying its performance. The changes in training loss and validation loss are shown in Fig. 12 (a), and the corresponding accuracies are also shown in Fig. 12 (b). Both training and validation losses decrease with the increase in the number of epochs, and converge to acceptable values after 60 epochs. The validation loss approaches the training loss, indicating good model fit to the training dataset. Notably, the validation loss remains a small margin above the training loss, suggesting that ETCN avoids overfitting. At the same time, training and validation accuracies quickly exceed 90%, and approach 100% with 30 training epochs, demonstrating its effectiveness in fault diagnosis.

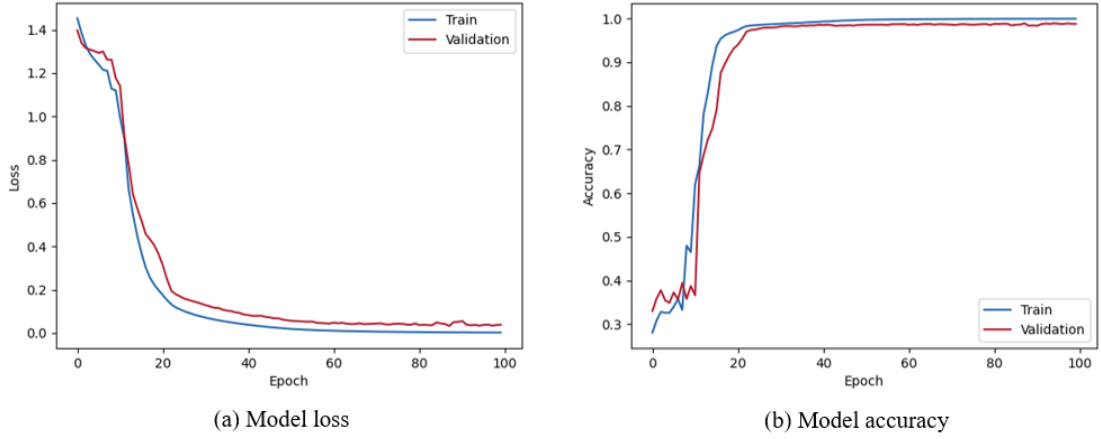

(a) Model loss  (b) Model accuracy

Fig. 12. Results of model training and validation.

## 4.2 Model testing and comparison

To evaluate the performance of ETCN and SSA, a separate test dataset is used after training and validation. For consistency, the same data split for training, validation, and testing across is applied across all models (including eight models detailed in Table 5). SSA is combined with each model and the entire program runs 20 times independently for each model configuration to reduce the impact of randomness. During the training process, a network structure similar to that of ETCN is used for each model. Here's a list of the different model types used. CNN possesses the ability to retain the inherent characteristics of time series data, while LSTM excels at capturing long-range temporal dependencies in time series data. Compared to LSTM, GRU has fewer parameters, leading to faster training time. TCN without SA mechanism and residual block is used to assess the intrinsic fault diagnosis ability of TCN. The combination of SA mechanism and TCN allows us to investigate the impact of residual block on model performance. Model6 without SA mechanism specifically explores the impact of SA mechanism on model performance. Model7 is designed to verify which type of residual block has a better effect on enhancing model performance. Finally, to demonstrate the superiority of ETCN, the results of Model8 are compared with it. The evaluation metrics introduced in Section 3.2 are used to assess and compare the fault diagnosis performance of all models.

Table 5: Results of evaluation metrics.

| Model | | Composition of model | Accuracy | Precision | Recall | F1 |
|---|---|---|---|---|---|---|
| ETCN | | TCN、SA、ResBlock2 | **0.9880** | **0.9898** | **0.9798** | **0.9844** |
| Comparative models | Model1 | CNN | 0.9254 | 0.9467 | 0.8783 | 0.9026 |
| | Model2 | LSTM | 0.9329 | 0.9325 | 0.9031 | 0.9154 |
| | Model3 | GRU | 0.9441 | 0.9541 | 0.9128 | 0.9289 |
| | Model4 | TCN | 0.9472 | 0.9491 | 0.9294 | 0.9204 |
| | Model5 | TCN、SA | 0.9798 | 0.9790 | 0.9671 | 0.9724 |
| | Model6 | TCN、Block2 | 0.9669 | 0.9751 | 0.9484 | 0.9589 |

| Model7 | TCN、SA、ResBlock1 | 0.9829 | 0.9803 | 0.9797 | 0.9799 |
| Model8 | GRU、SA、ResBlock2 | 0.9852 | 0.9873 | 0.9742 | 0.9801 |

Table 5 summarizes the comparison between comparative models and the suggested model employed. Although Model1-4 have some capability in handling time series data characteristics, and their fault diagnosis performance is relatively weaker compared with the other models evaluated. The comparison between Model6 and ETCN reveals that SA mechanism significantly improves the accuracy of fault diagnosis on time series data. Additionally, by comparing the diagnosis results of Model5, Model7 and ETCN, it can be seen that the effectiveness of residual block is highlighted in enhancing fault diagnosis ability and ResBlock2 appears to be more beneficial for improving diagnosis compared to ResBlock1. Since the diagnosis results of GRU and TCN have their own advantages and disadvantages separately, the control group of Model8 and ETCN is added. The experimental results convincingly demonstrate that ETCN achieves superior fault diagnosis results overall. In summary, ETCN effectively captures and retains the crucial time dependencies between historical observations in the time series data, leading to more accurate fault diagnosis.

To further verify the proposed method's performance, the confusion matrix for these models visualizes their ability to correctly diagnose different failure modes, as shown in Fig. 13. It can be observed that Model1-4 exhibit a high misdiagnosis rate for SGTR, most commonly misclassifying them as LOCA. This can be attributed to the similarity between certain aspects of these accidents in NPPs, particularly those causing damage to the primary circuit's pressure boundary. Looking at Fig. 13 (e), (f), and (h), we can see that models lacking SA mechanism achieve lower accuracy in diagnosing LOCA and SGTR. The strong feature extraction capabilities introduced by SA mechanism significantly improve the diagnosis accuracy for these faults, as shown in Fig. 13 (e) and (h). Furthermore, comparing Fig. 13 (e), (g), and (h), we observe that residual block (ResBlock2 specifically) offers additional improvement in fault diagnosis accuracy across various operating conditions. Finally, analyzing all subgraphs in Fig. 13, it's evident that ETCN surpasses other models in correctly diagnosing samples for each fault type, which demonstrates its superior feature extraction capabilities. Overall, the results confirm that combining SA mechanism and ResBlock2 within TCN architecture leads to a better fault diagnosis model.

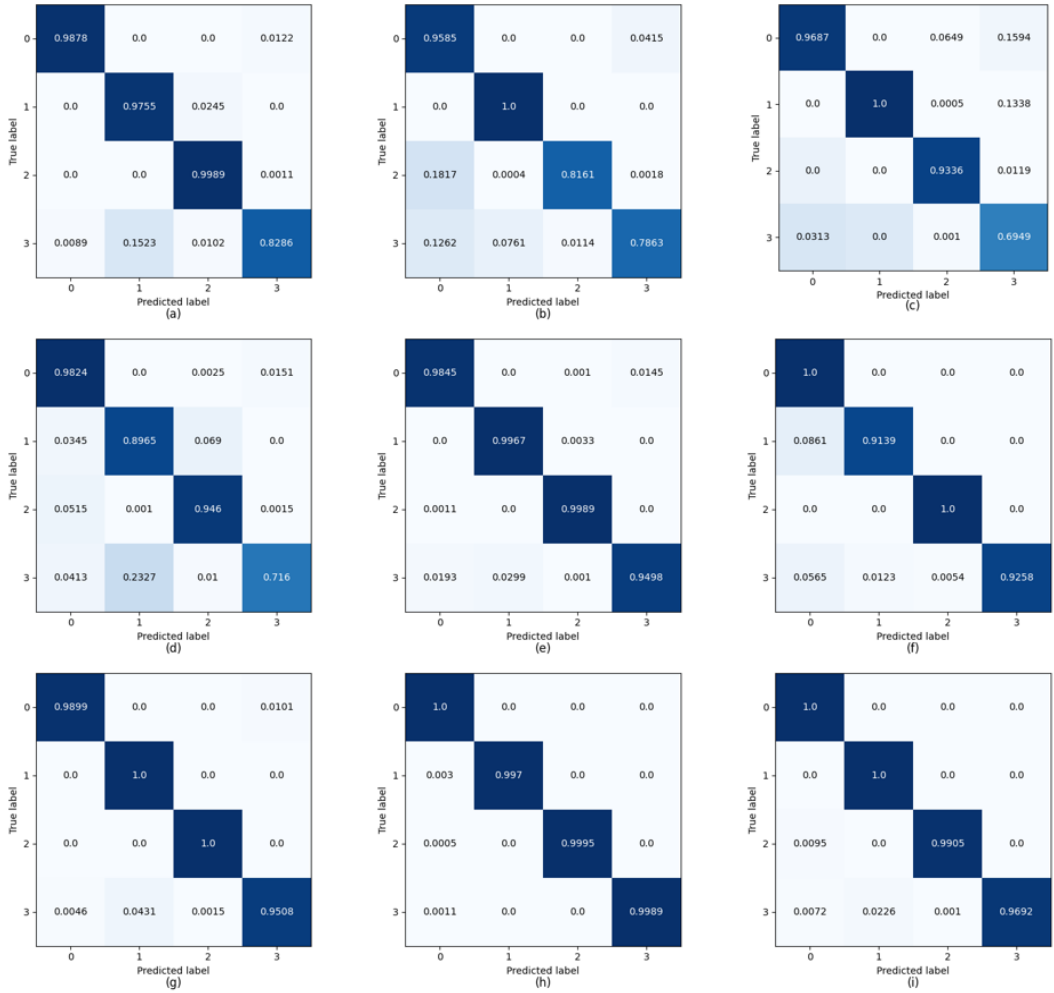

Fig. 13. Confusion matrix of (a) Model1, (b) Model2, (c) Model3,

(d) Model4, (e) Model5, (f) Model6, (g) Model7, (h) ETCN, (i) Model8.

To more vividly illustrate the feature extraction capability of ETCN, t-distributed stochastic neighbor embedding (t-SNE) is employed for dimensionality reduction visualization on feature extraction results. The t-SNE allows us to visualize high-dimensional data in a lower-dimensional space (typically two or three dimensions), making it easier to analyze the distribution of these features. Fig. 14 shows the feature extraction performance on the test dataset, where each color mark represents an operational condition. As can be seen from Fig. 14 (a), (b) and (c), the fault data points processed by individual modules like CNN, LSTM and GRU appear partially mixed, with blurred boundaries between different fault categories. Fig. 14 (d) and (e) depict the effect of SA mechanism. While TCN module individually extracts features, it can be observed that some different categories of data remain mixed together. However, after passing through SA mechanism module, the distribution of data for each category becomes significantly more distinct (Fig. 14 (e)). This highlights the role of SA mechanism in focusing on informative features for each fault type. Finally, after feature extraction through ResBlock2 module (Fig. 14 (h)), different types of faults form well-separated clusters with clear boundaries. The significant reduction in crossover situations and the emergence of distinct output feature distributions indicate that these features contain valuable information for fault classification. In essence, the four types of faults can now be effectively distinguished. The visualization further support ETCN's strong performance.

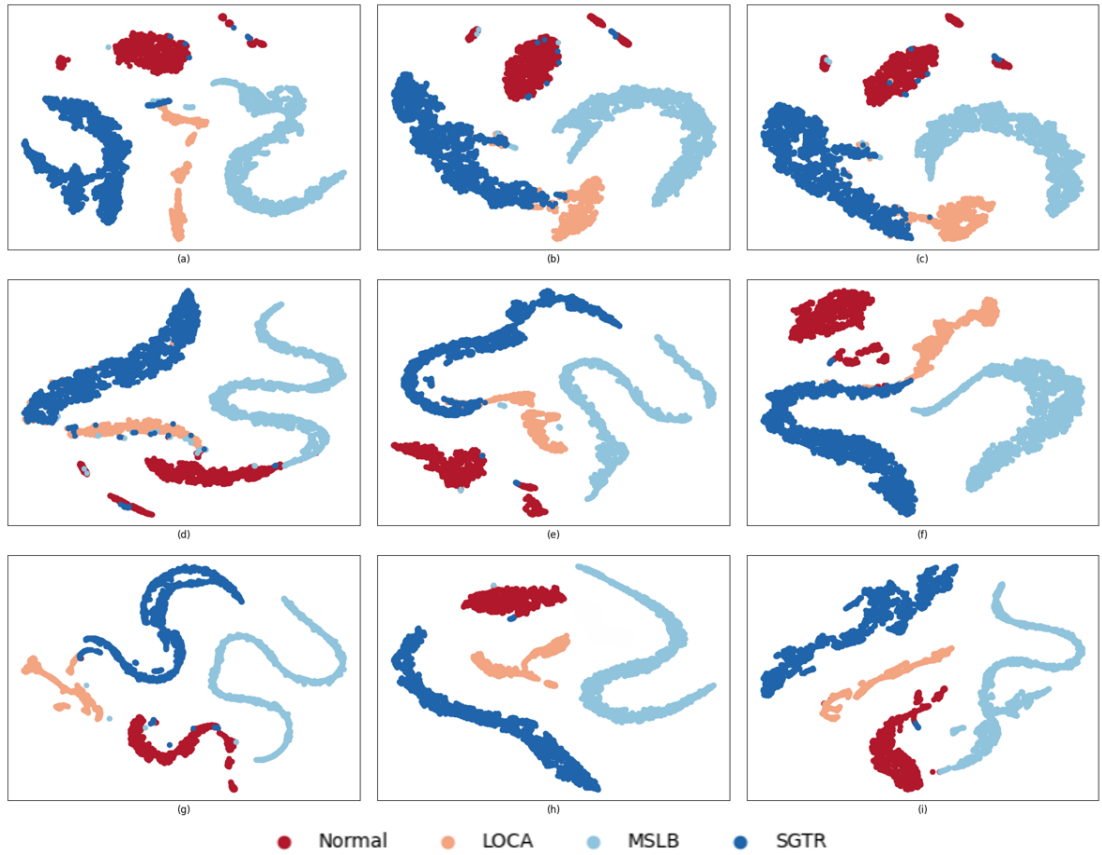

Fig. 14. Feature visualization of (a) Model1, (b) Model2, (c) Model3,

(d) Model4, (e) Model5, (f) Model6, (g) Model7, (h) ETCN, (i) Model8.

The results convincingly demonstrate that the proposed ETCN surpasses other methods in fault diagnosis performance. This success can be attributed to the synergistic combination of TCN, SA mechanism and residual block. TCN effectively extracts temporal features from multivariate time series data, allowing for the capture of subtle temporal relationships between closely related variables. SA mechanism dynamically adjusts attention weights based on input temporal features. This enables the extraction of internal autocorrelation within time series data, leading to a better understanding of the temporal dependencies within these features. Residual block improves accuracy by increasing its depth. By combining these modules, ETCN achieves superior fault diagnosis ability by effectively capturing the temporal characteristics and relationships within NPP data. This improved capability has significant implications for NPPs that rely on accurate and timely fault diagnosis in time series data, potentially leading to enhanced safety, efficiency, and cost-effectiveness.

# 5 Conclusion

NPPs operate under stringent safety and reliability requirements. Timely and accurate fault diagnosis is crucial for the safety and continuous operation of NPPs. Given the advancements in AI, neural networks have emerged as a promising approach for FDD system in NPPs, owing to superior feature extraction abilities and autonomous learning potential. To address the challenges and enhance the flexibility and reliability of FDD system, this paper proposes an intelligent fault diagnosis method based on ETCN-SSA combined algorithm. The results demonstrate that ETCN optimized with SSA achieves superior fault diagnosis accuracy compared with others. From the above discussion, the following conclusions can be drawn:

(1) The selection of sliding window parameters is crucial. Appropriate sliding window width and step size help the model learn the temporal relationships between preceding and subsequent data points, better reflecting the operating status of NPPs.

(2) SSA plays a vital role in optimizing ETCN's hyperparameters. It aims to find the hyperparameter configuration that leads to the best possible model performance and effectively improves the results of model fault diagnosis.

(3) Complex hybrid models, like the proposed ETCN, have demonstrably superior diagnostic capabilities compared to the models relying on single module. According to experimental results, the combination of TCN, SA mechanism and residual block can achieve a significant performance improvement in NPP fault diagnosis.

(4) The proposed ETCN demonstrates superior fault diagnosis performance. Compared with other hybrid models, ETCN has a more outstanding ability to extract features of nuclear power units.

In summary, the proposed ETCN is a promising fault diagnosis method that can effectively diagnose NPP faults under different operational conditions. Although these valuable improvements have been made and the expected results have been obtained, there is still some work for further exploration. Future work will focus on improving the model's generalization ability to handle faults at different power levels and achieve the diagnosis of unknown faults. Besides, while this study employed Gaussian noise for data augmentation, real-world sensor data might exhibit different noise characteristics. Future research will investigate the impact of non-Gaussian noise on the model's performance and explore potential mitigation strategies. By addressing these limitations and continuing in-depth exploration, we aim to further improve ETCN and propose even more effective solutions for maintaining safety and reliability of NPP operation.

**Declaration of competing interest**

The authors declare that there is no conflict of interest.

**Acknowledgement**

This work was supported by the Science and Technology Project of Fujian Province (No. 2022H0004).